\documentclass[twocolumn]{article}
\usepackage[utf8]{inputenc}
\usepackage{amsmath}
\usepackage{amsfonts}
\usepackage{amssymb}
\usepackage{mathrsfs}
\usepackage{multirow}
\usepackage{balance}
\usepackage{tcolorbox}

\usepackage{float}
\usepackage{algorithm}
\usepackage{algorithmic}

\usepackage{blindtext}

\usepackage{MnSymbol}

\usepackage[cal=boondox,scr=boondoxo]{mathalfa}

\usepackage{float}
\usepackage{fancybox,graphicx}
\usepackage{subfig}
\usepackage{caption}
\usepackage{color}
\usepackage{authblk}

\usepackage[colorlinks]{hyperref}
\usepackage{hyperref}

\newcommand{\doi}[1]{{doi:~\href{https://doi.org/#1}{\nolinkurl{#1}}}\rmFullStop}

\newcommand*{\rmFullStop}{\rmifnextchar{.}{}{}}

\makeatletter
\newcommand{\rmifnextchar}[3]{%
  \begingroup
  \ltx@LocToksA{\endgroup#2}%
  \ltx@LocToksB{\endgroup#3}%
  \ltx@ifnextchar{#1}{%
    \def\next{\the\ltx@LocToksA}%
    \afterassignment\next
    \let\scratch= %
  }{%
    \the\ltx@LocToksB
  }%
}
\makeatother %doi command

\usepackage{accents}
\usepackage[titletoc,title]{appendix}
\usepackage{cite}

\usepackage{mathtools}

\usepackage[top=2in, bottom=1.5in, left=1in, right=1in]{geometry}

\usepackage{stackengine}

\title{ChartComplete: A Taxonomy-based Inclusive Chart Dataset} 
 
\author{Ahmad Mustapha}
\author{Charbel Toumieh}
\author{Mariette Awad}

\affil{American University of Beirut, Lebanon}
\begin{document}

\maketitle

\begin{abstract}

With advancements in deep learning (DL) and computer vision techniques, the field of chart understanding is evolving rapidly. In particular, multi-modal large language models (MLLMs) are proving to be efficient and accurate in understanding charts. To accurately measure the performance of MLLMs, the research community has developed multiple datasets to serve as benchmarks. By examining these datasets, we found that they are all limited to a small set of chart types. To bridge this gap, we propose the ChartComplete dataset. The dataset is based on a chart taxonomy borrowed from the visualisation community, and it covers thirty different chart types. The dataset is a collection of classified chart images and doesn't include a learning signal. We present the ChartComplete dataset as is to the community to build upon it.

\end{abstract}

\section*{Background} 

Artificial intelligence systems are rapidly becoming smarter and capable of handling more complex tasks. One of these complex tasks is understanding charts. In particular, ChartQA is a research domain where machines are designed to have a visual understanding of chart elements. In ChartQA, a machine is handed a question and a chart, and the machine is supposed to answer the question about this image. To be successful in answering the question, the machine is required to perform a complex task that is composed of computer vision, natural language processing, and domain knowledge, all in one shot.

A handful of benchmarks have been proposed to test machine learning models' capabilities in handling ChartQA questions. However, the majority of these benchmarks are limited to a few types of charts, such as bar, pie, line, and scatter charts. To truly test a model's capabilities in ChartQA, the benchmarks should be more inclusive and include more types of chart. Ideally, models should support common chart types. To bridge this gap, we present the ChartComplete dataset. ChartComplete includes 30 different types of chart, ranging from the common bar charts to the less common charts, such as parallel coordinates charts, including special charts such as box and whisker charts and choropleth map charts. Unlike existing benchmarks, ChartComplete was based on a chart taxonomy that was borrowed from the visualization taxonomy.

\begin{figure}
    \centering
    \includegraphics[width=1\linewidth]{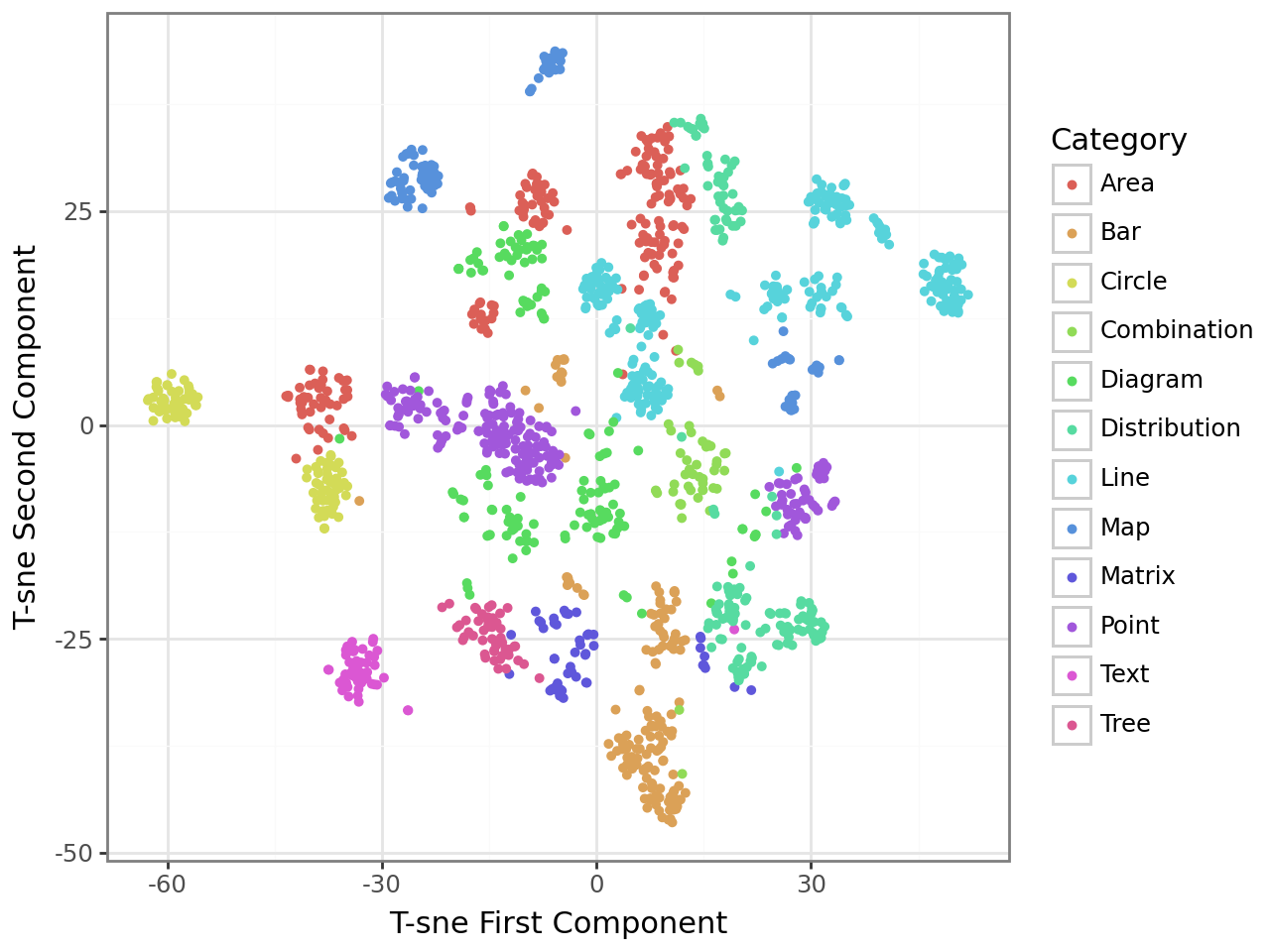}
    \caption{The visual feature space of ChartComplete images. The features are extracted using Google VIT. The space is projected using T-SNE.}
    \label{fig:features-distribution}
\end{figure}

One of the first works on the topic was published under the name FigureSeer \cite{FigureSeer}. Following the trace of FigureSeer but with scalability in mind, FigureQA \cite{Kahou2017} was proposed. FigureQA is the first large-scale dataset that addresses the ChartQA question. Both the data for the charts and the charts themselves were generated by the authors. To solve the problem of unnatural data labels, PlotQA \cite{Methani2019} was proposed. Unlike FigureQA, the chart data were collected from different sources. The charts themselves were generated. As a sequel in this line of research, ChartQA \cite{masry2022} was proposed. ChartQA collected charts and their data from online resources, making it the first dataset to have real-life-looking charts with real-life data. Other data sets include Chart-to-text \cite{chart2text}, OpenCQA \cite{opencqa}, and ChartFC \cite{Akhtar2023}. 

All of these data sets and benchmarks presented above addressed only a handful of types of charts. Those charts were mainly bar charts, horizontal and vertical, grouped and stacked, line charts, and pie charts. Table \ref{tab:existing-benchmarks} summarizes the chart types used by different benchmarks.
%Grammarly

\begin{table}[h]
\caption{Summary of the chart types covered by existing datasets}
\centering
\begin{tabular}{|l|p{5cm}|}
\hline
Dataset & Supported Charts \\
\hline
FigureSeer & Graph Plot, Flow Chart, Scatter Plot, Bar Plot\\
\hline
FigureQA & Line Plot, Dot-Line Plot, Vertical Bar Plot, Horizontal Bar Plot, Pie Plot \\
\hline
PlotQA & Bar Plot, Scatter Plot, Line Plot \\
\hline
ChartQA & Horizontal Bar Plot, Vertical Bar Plot, Stacked Bar Plot, Pie Chart, Line Plot \\
\hline
Chart-to-Text & Bar Plot, Grouped Bar Plot, Stacked Bar Plot, Line Plot, Area Plot, Scatter Plot, Pie Plot \\
\hline
OpenCQA & Bar Plot, Grouped Bar Plot, Stacked Bar Plot, Line Plot, Area Plot, Scatter Plot, Pie Plot \\
\hline
ChartFC & Bar Plot\\

\hline
\end{tabular}
\label{tab:existing-benchmarks}
\end{table}

\begin{figure}
    \centering
    \includegraphics[width=0.9\linewidth]{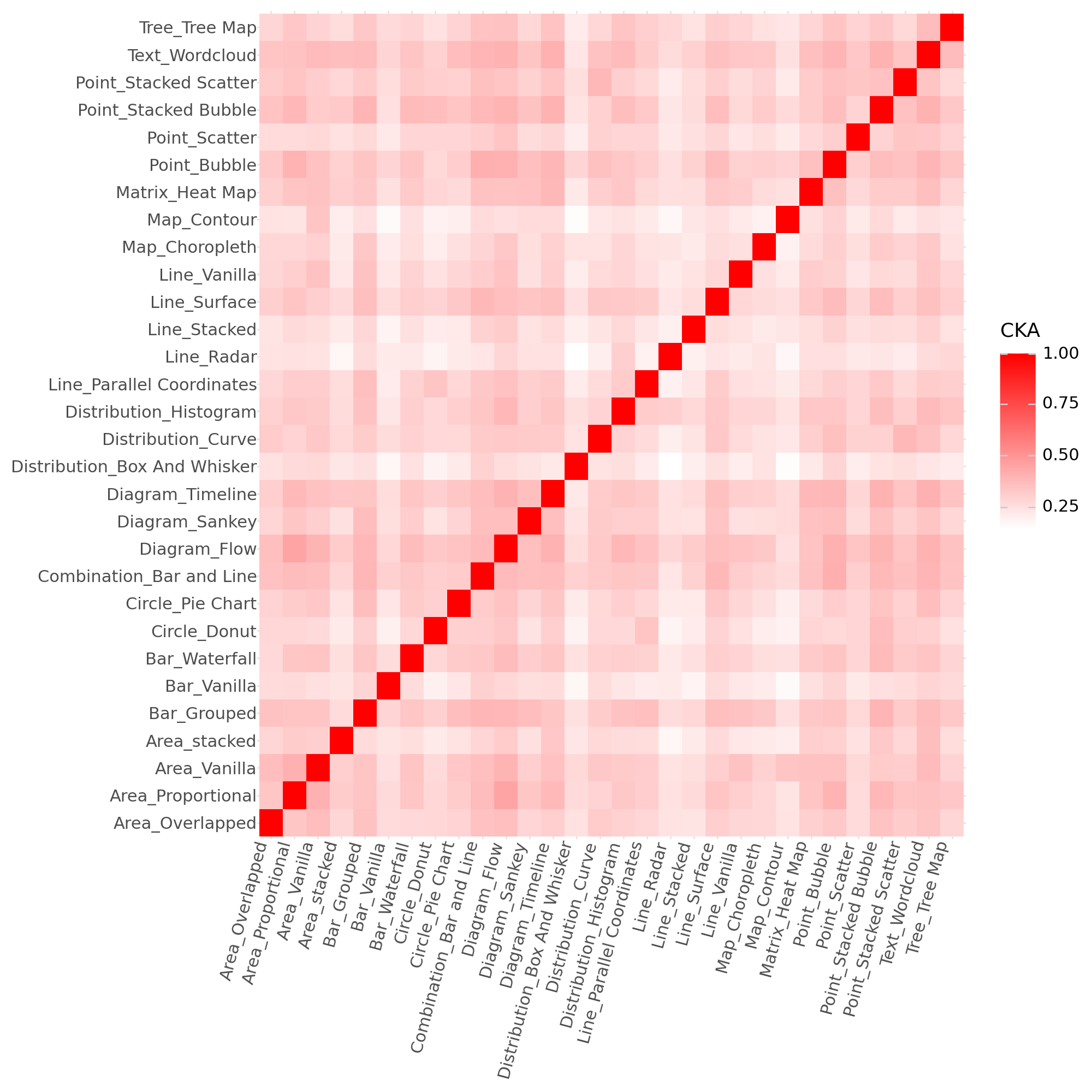}
    \caption{The Centered Kernel Alignment (CKA) values between different chart types features. The features are extracted using Google ViT. The higher the CKA value the more the features are similar.}
    \label{fig:cka}
\end{figure}

A current limitation of ChartComplete is that it serves solely as a dataset of classified chart images. It presently lacks a direct training signal, such as those used for summarization or question answering tasks. We plan to address this in future work and actively encourage the community to contribute to its development in these areas. 

\section*{Collection Methods and Design}

The collection process of the ChartComplete dataset was based on a modified version of Borkin's taxonomy of chart types \cite{Borkin}. The taxonomy has categories of charts and terms, or chart types in each category. We have thirty different chart types. For each chart type, we collected fifty images. The collection process was partially manual and partially automated, utilising scraping. Table \ref{tab:scraped-collected} shows the counts and the percentage of manually collected images vs scraped ones.

% Grammarly

\begin{table}[h]
\caption{Summary of collected and scraped charts images in the  ChartComplete dataset}
\centering
\begin{tabular}{|l|r|r|}
\hline
Type      & Count & Percentage (\%) \\
\hline
Collected & 951   & 63.4            \\ \hline
Scraped   & 549   & 36.6            \\
\hline
\end{tabular}
\label{tab:scraped-collected}
\end{table}

Table \ref{tab:borkin} presents the taxonomy on which we based our work. It is a modified version of Borkin's taxonomy \cite{Borkin}.

As aforementioned, we introduced some changes to the original Borkin taxonomy. First, we introduced the "stacked" variant of different charts. We define the "stacked" chart as a chart that represents several data columns in the same plot. Several lines in a line chart, for example. They usually have several legends. The naming convention was inspired by the common "stacked bar chart". In the literature, such charts are differentiated from simpler charts by using the simple/complex prefixes \cite{opencqa}. For clarity, we decided to use the vanilla/stacked prefixes. We introduced the stacked class to four chart types: area, line, scatter, and bubble charts. The charts that represent only one data column are referred to as "vanilla" charts. Second, we removed from the taxonomy the less common chart types like the Venn diagram, the circular bar chart, or the contour line chart. Third, we added combination charts, like the bar and line charts. Table \ref{tab:borkin} represents the entire taxonomy. 
% Grammarly

\begin{table}[]
    \centering
    \caption{The modified Borkin taxonomy used to build ChartComplete}
    \label{tab:borkin}
    \begin{tabular}{|l|p{5cm}|}
    \hline
    \textbf{Category} & \textbf{Terms} \\
    \hline
    Area & Vanilla, Overlapped, Stacked, Proportional\\
    \hline
    Bar & Vanilla, Grouped, Waterfall\\
    \hline
    Circle & Donut, Pie\\
    \hline
    Diagram & Flow, Sankey, Timeline \\
    \hline
    Distribution & Curve, Histogram, Box and Whisker \\
    \hline
    Matrix & Heat Map \\
    \hline
    Line & Vanilla, Stacked, Radar, Surface, Parallel Coordinates \\
    \hline
    Map & Choropleth, Contour \\
    \hline
    Point & Scatter, Stacked Scatter, Bubble, Stacked Bubble\\
    \hline
    Text & Word Cloud \\
    \hline
    Tree & Tree Map \\
    \hline
    Combination & Bar and Line Chart \\
    \hline
    \end{tabular}
\end{table}

To collect the ChartComplete dataset we utilized the following sources:

\begin{itemize}
    \item \textbf{Source 1:} Statista \footnote{www.statista.com/chartoftheday/} website, which is a global data and business intelligence platform with an extensive collection of statistics, reports, and insights on over 80,000 topics from 22,500 sources in 170 industries. We scraped all the charts listed on the chart of the day webpage on the 27th of January 2025.

    \item \textbf{Source 2:} Our World in Data \footnote{www.ourworldindata.org} website, which is a project of the Global Change Data Lab, a non-profit organisation based in the United Kingdom. We scraped all the charts in the "Browse By Topic" menu item under the "Charts" menu item for each topic if available.

    \item \textbf{Source 3:} Different online sources. A sizable part of the dataset was not scraped but was collected. This is due to the lack of a large collection of charts provided in one place for some chart types. 
    
\end{itemize}

The number of scraped charts was large, 12,635 and 4,113 from Statista and OurWorldInData, respectively, and the type of charts was not provided. Our goal was to collect 50 chart images for each chart type in the taxonomy. To extract the required charts from the entire collection of scraped charts, we followed algorithm \ref{alg:filter}.

We first extracted the image features of all the scraped images using a Google ViT \footnote{huggingface.co/google/vit-base-patch16-224}. Then we build an index of images using FAISS \cite{johnson2019billion}. For each chart type, we sampled a random image until we got an image of the required type. We then get the 100 nearest neighbours of the selected image. We manually filter the correct images out of the 100 images. We repeat the process until we get 50 images for each type. Note that the scraped images covered only a handful of chart types. The other charts are completed to 50 through manual collection. 
%Grammarly

\begin{algorithm}
\caption{Chart Images Selection}
\label{alg:chart_selection}
\begin{algorithmic}[1]
\REQUIRE Scraped images dataset $D_0$
\REQUIRE Google vision transformer $G_{ViT}$
\REQUIRE Number of nearest neighbors $K = 100$
\REQUIRE Number of images per chart type $C = 50$
\STATE Apply $G_{ViT}$ on $D_0$ to get $D_{ViT}$
\STATE Build a FAISS index $I$ from $D_{ViT}$
\FOR{each chart type $T$}
    \STATE \textbf{Let} $D_T \leftarrow$ Selected images of chart type $T$ 
    \STATE Initialize $D_T \leftarrow \emptyset$ 
    \STATE Randomly sample image $i$ from $D_0$
    \WHILE{ $|D_T| < C$}
        \IF{$i$ is of type $T$}
            \STATE \textbf{Let} $f_i$ be the feature vector of $i$
            \STATE Use $I$ to find $K$ nearest neighbors of $f_i$
            \STATE Manually inspect the $k$ nearest images
            \STATE Add correct images to $D_T$
        \ENDIF
    \ENDWHILE
\ENDFOR
\label{alg:filter}
\end{algorithmic}
\end{algorithm}

\begin{figure}
    \centering
    \includegraphics[width=0.8\linewidth]{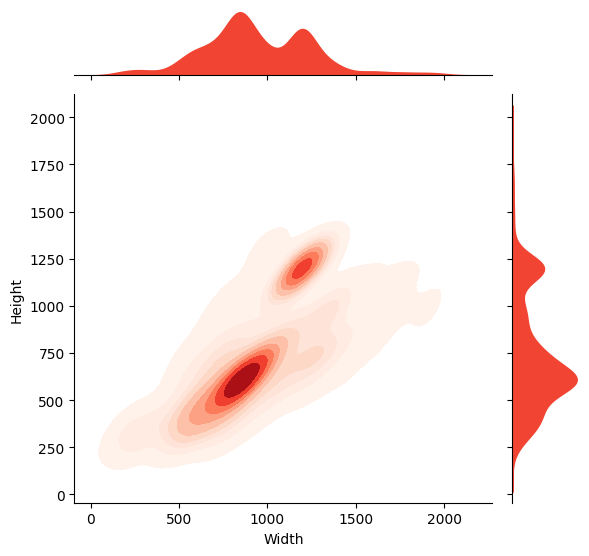}
    \caption{The distribution of the image sizes in the ChartComplete dataset}
    \label{fig:sizes}
\end{figure}

\section*{Validation and Quality} 

The scraped images only covered a handful of chart types. For the other not-covered chart types, we launched a collection campaign. The first round of collection resulted in low-quality images that didn't meet quality requirements. For this, we developed a collection guideline. The subsequent collection iteration led to images of higher quality.

The guidelines are as following:

\begin{itemize}
    \item \textbf{Image Quality}. The images collected for the charts should be of high quality. The size/content ratio should be suitable so that if we zoom into the image, it doesn’t get blurry. The font should be readable, and the image should not contain any glitches. The minimum allowed dimension of an image is 300 pixels. Ensure readability by avoiding text that is too small or light-colored.
    \item \textbf{Content Quality}. The image content should also be of high quality, corresponding to a real-life scenario that serves the objective of providing real information. If the image content is illustrative or serves a visual aspect, it shouldn’t pass the selection process. 
    \item \textbf{Complete Information}. The images of the charts should provide complete information. In other words, it has a clear title and clear labels for the axis, if any. Axes must be labeled with clear units, for example, “Revenue (\$M)” or “Years”. Legends should be available if required by the chart type. 
    \item \textbf{No Prior Knowledge}. The images should not require prior visual knowledge. For example, using companies' logos as legends or labels.
    \item \textbf{Complexity}. The images should not be overly complex or fancy. The images should strike a balance between simplicity and complexity. 
    \item \textbf{Image Type}. The image should be in one of the following image formats: JPEG or PNG. 
\end{itemize}

The guidelines emphasized collecting high-quality, informative chart images suitable for real-life scenarios. Images must be clear, with readable fonts, proper resolution, and free of glitches. Charts should present complete information, including titles, labeled axes with units and ranges, and legends when needed. Only textual labels are allowed; logos or symbols requiring prior knowledge are not permitted. The visuals should strike a balance between simplicity and complexity, avoiding overly fancy or illustrative styles. Acceptable formats for submission are JPEG and PNG.

The collection process was iterative and used version control. The guidelines were explained to the campaign participants. The collection tasks were iteratively distributed. After the collection is finished, the collected images are pushed to the version control repository. Before merging to the main branch, a quality check was performed by the authors to make sure the collected images meet the guidelines requirements. Images that didn't meet the requirements were removed. The process is repeated until the target of 50 high quality images per chart type is met.

After collecting the entire dataset, another quality check routine was performed to check the entire dataset for alignment with the guidelines. Also, duplicated images were removed. And images that have "No" as an answer to the following question, "Can we ask meaningful questions about the chart?" were also removed.
% Grammarly

% \begin{figure}
% \centerline{\includegraphics[width=3.5in]{fig1}}
% \caption{Magnetization as a function of applied field. Note that ``FIG.'' 
% is abbreviated. There is a period after the figure number, followed by two 
% spaces. It is good practice to explain the significance of the figure in the 
% caption.\label{fig1}}
% \end{figure}

% \begin{table}
% \caption{Units for Magnetic Properties}
% \label{table}
% \setlength{\tabcolsep}{3pt}
% \begin{tabular}{|p{25pt}|p{75pt}|p{115pt}|}
% \hline
% Symbol& 
% Quantity& 
% Conversion from Gaussian and \par CGS EMU to SI $^{\mathrm{a}}$ \\
% \hline
% $\Phi $& 
% magnetic flux& 
% 1 Mx $\to  10^{-8}$ Wb $= 10^{-8}$ V$\cdot $s \\
% $B$& 
% magnetic flux density, \par magnetic induction& 
% 1 G $\to  10^{-4}$ T $= 10^{-4}$ Wb/m$^{2}$ \\
% $H$& 
% magnetic field strength& 
% 1 Oe $\to  10^{3}/(4\pi )$ A/m \\
% \hline
% \end{tabular}
% \label{tab1}
% \end{table}

\section*{Records and Storage} 

ChartComplete is stored as a directory. The main directory contains 12 root directories representing the category of chart types. In each directory, we have several directories, each representing the chart types. In each type directory, we have 50 chart images. The keywords "collected" and "scraped" are written as part of the images' names, representing the method used for collecting each image.

% Grammarly

\balance
\section*{Insights and Notes} 

\begin{figure}
    \centering
    \includegraphics[width=1\linewidth]{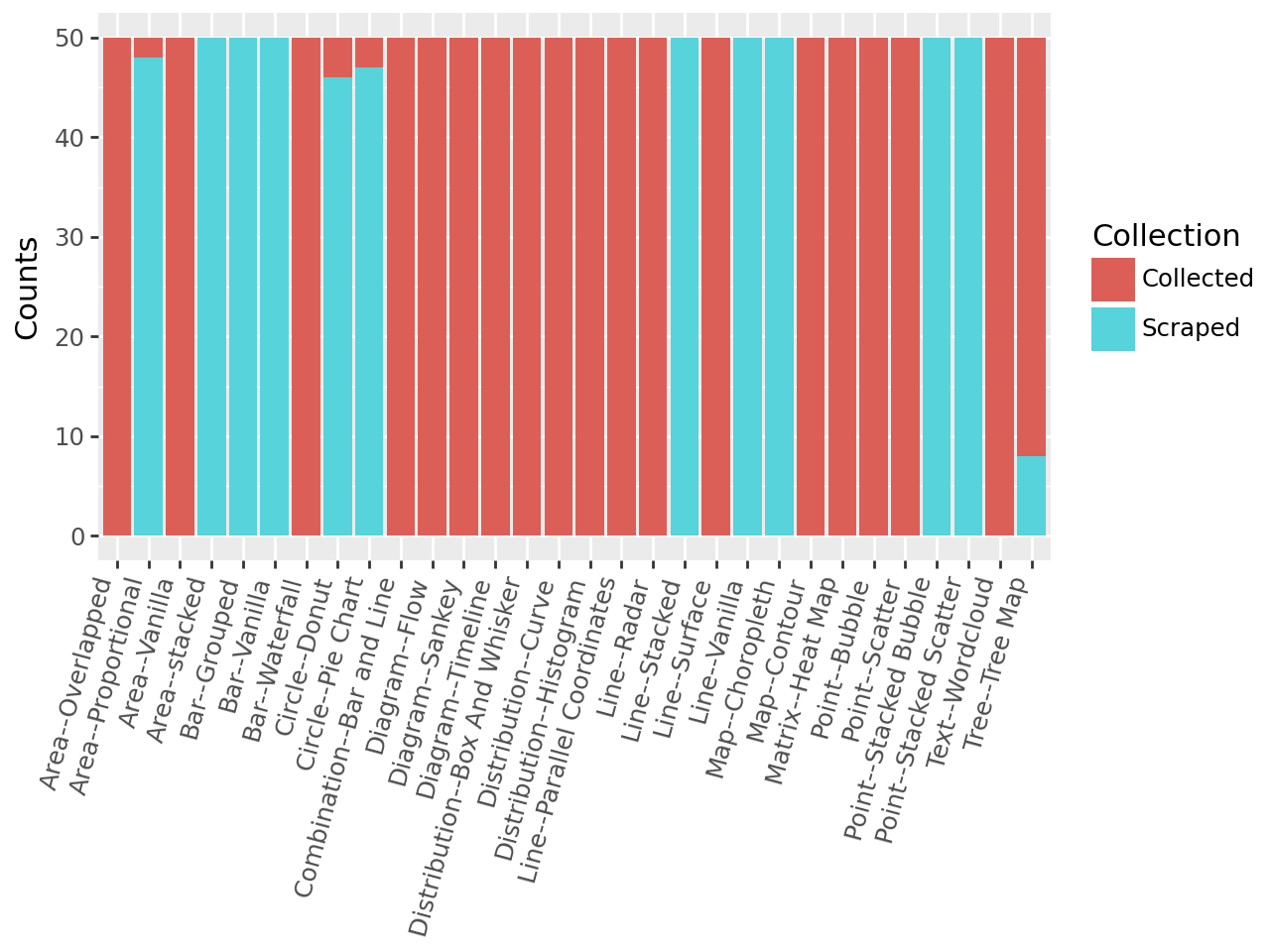}
    \caption{The distribution of collection methods per chart type.}
    \label{fig:collection-methods}
\end{figure}

Figure \ref{fig:collection-methods} shows the distribution of collection methods per chart type. We have 18 chart types that were purely collected manually, 8 chart types were purely scraped, and 4 chart types were both scraped and manually collected.

Figure \ref{fig:sizes} shows the distribution of ChartComplete image sizes. The distribution is not uniform as we didn't included the image size or shape in our collection guidelines. The only guideline was that the image should be of size greater than 300 pixels on both dimensions. The figure shows that in general the image resolution of ChartComplete dataset is large.

Figure \ref{fig:features-distribution} shows the visual features of ChartComplete chart images distributed in a two-dimensional space using T-SNE \cite{tsne}. The visual features were extracted using Google VIT \footnote{huggingface.co/google/vit-base-patch16-224}. The space is segmented, which points out to both the diversity and the clusterization of the visual features and the model ability to capture relevant discriminative features. The space still has notable imperfections. For instance, some area charts appear clustered near circle charts, while other area chart groups are located close to diagram charts. This suggests that the dataset would benefit from a more specialized feature extraction model. 

To further study the similarity between the charts from the visual perspective we computed the Centered Kernel Alignment (CKA) metric \cite{cka} between all chart types. CKA is a similarity metric used to compare the representations (i.e., features or embeddings) learned by deep learning models. Figure \ref{fig:cka} shows the heatmap that represents the similarity between different chart types from visual perspective. The figure shows that all chart types are different from each other feature wise.

% Grammarly

\section*{Source Code} 

The self-contained source code that includes the charts collection is available on Github \href{https://github.com/AI-DSCHubAUB/ChartComplete-Dataset}{(AI-DSCHubAUB/ChartComplete-Dataset)}. The dataset is distributed under the CC BY license.

% Grammarly

\section*{Acknoledgments}

Special thanks to the Maroun Semaan Faculty of
Engineering and Architecture (MSFEA) at the American University of Beirut (AUB) for providing the essential infrastructure needed to conduct
this study.

A.M developed the research idea and carried out implementation. C.T contributed to data analysis and to the manual data collection campaign. M.A reviewed the manuscript and gave feedback. We sincerely thank Ahmad Hlayhel, Ali Slaman, Joud Senan, and Kassem El Moussawi for their significant contributions to the data collection process.

This work was funded by the AI, Data Science, and Computing Hub (AI-DSC) at the American University of Beirut (AUB).

The article authors have declared no conflicts of interest.

% Grammarly

\bibliographystyle{IEEEtran}
\bibliography{biblio}

\end{document}